\documentclass{article}

\usepackage{microtype}
\usepackage{graphicx}
\usepackage{subcaption}
\usepackage{booktabs} 

\usepackage{hyperref}

\usepackage[preprint]{icml2026}

\usepackage{amsmath}
\usepackage{amssymb}
\usepackage{mathtools}
\usepackage{amsthm}
\usepackage{multirow}

\usepackage[capitalize,noabbrev]{cleveref}

\theoremstyle{plain}

\theoremstyle{definition}

\theoremstyle{remark}

\usepackage[textsize=tiny]{todonotes}
\usepackage{enumitem}

\icmltitlerunning{Submission for ICML 2026}

\begin{document}

\twocolumn[
  \icmltitle{TADS: Task-Aware Data Selection for Multi-Task Multimodal Pre-Training}




  \begin{icmlauthorlist}
    \icmlauthor{Guanjie Cheng}{ZJUSE}
    \icmlauthor{Boyi Li}{NEUCS}
    \icmlauthor{Lingyu Sun}{NNUCS}
    \icmlauthor{Mengying Zhu}{ZJUSE}
    \icmlauthor{Yangyang Wu}{ZJUSE}
    \icmlauthor{Xinkui Zhao}{ZJUSE}
    \icmlauthor{Shuiguang Deng}{ZJUCS}
  \end{icmlauthorlist}

  \icmlaffiliation{ZJUSE}{School of Software Technology, Zhejiang University}
  \icmlaffiliation{NEUCS}{School of Computer Science and Engineering, Northeastern University}
  \icmlaffiliation{NNUCS}{School of Computer and Electronic Information, Nanjing Normal University}
  \icmlaffiliation{ZJUCS}{School of Computer Science and Technology, Zhejiang University}

  \icmlcorrespondingauthor{Shuiguang Deng}{dengsg@zju.edu.cn}


]



\printAffiliationsAndNotice{}  

\begin{abstract}
Large-scale multimodal pre-trained models like CLIP rely heavily on high-quality training data, yet raw web-crawled datasets are often noisy, misaligned, and redundant, leading to inefficient training and suboptimal generalization. Existing data selection methods are either heuristic-based, suffering from bias and limited diversity, or data-driven but task-agnostic, failing to optimize for multi-task scenarios. To address these gaps, we introduce TADS (\textbf{T}ask-\textbf{A}ware \textbf{D}ata \textbf{S}election), a novel framework for multi-task multimodal pre-training that integrates \textit{Intrinsic Quality}, \textit{Task Relevance}, and \textit{Distributional Diversity} into a learnable value function. TADS employs a comprehensive quality assessment system with unimodal and cross-modal operators, quantifies task relevance via interpretable similarity vectors, and optimizes diversity through cluster-based weighting. A feedback-driven meta-learning mechanism adaptively refines the selection strategy based on proxy model performance across multiple downstream tasks. Experiments on CC12M demonstrate that TADS achieves superior zero-shot performance on benchmarks like ImageNet, CIFAR-100, MS-COCO, and Flickr30K, using only 36\% of the data while outperforming baselines by an average of 1.0\%. This highlights that TADS significantly enhances data efficiency by curating a high-utility subset that yields a much higher performance ceiling within the same computational constraints.
\end{abstract}
\begin{figure}[t]
    \centering
    \includegraphics[width=1\linewidth]{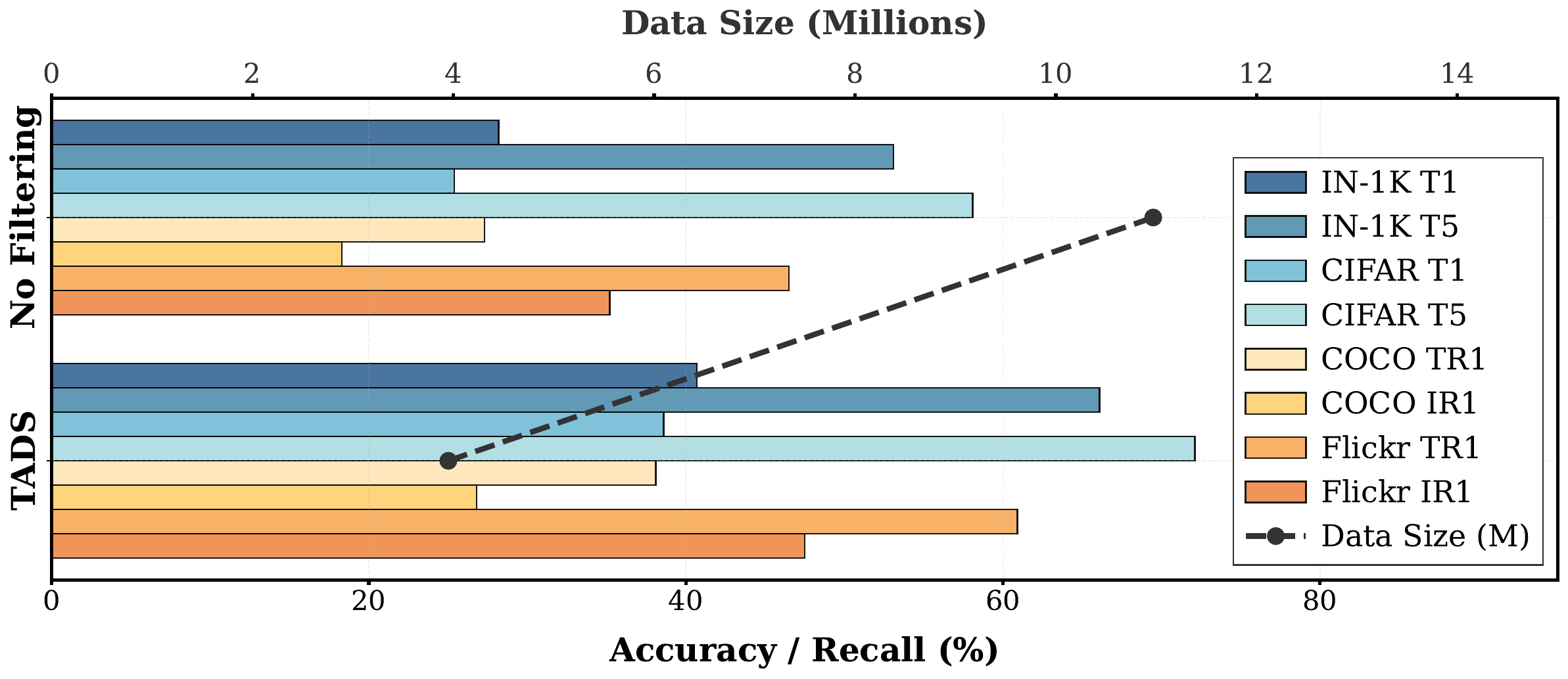}
    \caption{Motivation: Quality outweighs Quantity. We compare the zero-shot performance of CLIP models trained on the full, uncurated CC12M dataset versus a compact subset selected by TADS. As illustrated, although TADS reduces the training data scale by 64\%, it consistently achieves superior accuracy and recall across all 8 downstream benchmarks. This foundation experiment demonstrates that task-irrelevant noise, rather than data scarcity, is the primary bottleneck in multimodal pre-training, validating the necessity of our proposed task-aware selection framework.}
    \label{fig1_introduction}
\end{figure}

\section{Introduction}
In recent years, large-scale multimodal pre-trained models such as CLIP \cite{radford2021learning}, BLIP-2 \cite{li2023blip}, SigLIP-2 \cite{tschannen2025siglip}, and SAIL \cite{lei2025scalability} have emerged as the cornerstone of cross-modal understanding tasks, demonstrating exceptional generalization capabilities in scenarios including zero-shot classification \cite{atabuzzaman2025zero,tang2025language}, image-text representation learning \cite{guo2025mmrl,kim2025context}, and in-context learning \cite{doveh2024towards,chen2025provoking}. However, the performance of these foundation models depends critically on the quality and scale of the data that drives them \cite{hoffmann2022training,xu2025quality,kanjula2025understanding}. In real-world pre-training, relying on raw web-harvested datasets—inevitably plagued by dense noise, semantic misalignment, high redundancy \cite{abbas2023semdedup} and weak task relevance —imposes prohibitive computational costs and often compromises downstream generalization. 
Consequently, the issue of data selection—identifying high-value subsets from massive, noisy candidate pools—has become a core bottleneck impeding the advancement of efficient and robust multimodal learning. Figure~\ref{fig1_introduction} illustrates our motivation: task-aware data selection outperforms naive scaling of noisy data.

Existing approaches to multimodal data selection generally fall into heuristic-based or data-driven paradigms. Heuristic methods, exemplified by early filtering strategies in LAION \cite{schuhmann2021laion,schuhmann2022laion}, rely on rigid manual rules or single-dimensional indicators such as CLIP cosine similarity \cite{hessel2021clipscore,gadre2023datacomp}. These approaches, however, are limited by their over-reliance on human priors: first, the predefined logic fails to capture non-linear correlations across multiple quality dimensions; second, they lack systematic and comprehensive operator designs (i.e., scoring or evaluation functions that quantify sample quality across multiple criteria) for fine-grained assessment; third, rigid thresholds tend to favor dominant patterns, inadvertently filtering out long-tail samples and constraining dataset diversity \cite{mahmoud2024sieve}. Conversely, data-driven methods \cite{fang2023data,xu2025quality} aim to learn quality patterns but primarily estimate \emph{general data quality}, ignoring specific \emph{task relevance}. While recent task-aware methods \cite{shechter2025filter} introduce feedback signals, they predominantly focus on single-task optimization through implicit modeling. Crucially, current paradigms lack a mechanism for the collaborative optimization of multiple tasks, failing to resolve conflicts where samples beneficial for one task (e.g., OCR-heavy data) might degrade performance on another (e.g., zero-shot classification).

Beyond individual methodological limitations, three fundamental gaps hinder effective multi-task multimodal pre-training. First, quality assessment remains incomplete and fragmented. Single-indicator methods \cite{gadre2023datacomp} address only coarse defects, whereas multi-dimensional approaches often lack systematic operator designs. For example, global alignment metrics\cite{hessel2021clipscore} fail to capture local object-level mismatches \cite{liu2024grounding,li2022grounded}, and existing cross-modal operators rarely account for the complementary relationship between OCR-derived text and associated captions. Second, task awareness is insufficiently modeled in multi-task settings. Most methods either ignore task relevance altogether or optimize for a single downstream task, making it difficult to balance performance trade-offs across heterogeneous tasks. Although recent works \cite{wang2024cliploss,kim2024hype} attempt to quantify task adaptability through visual similarity, they lack mechanisms for collaborative optimization across multiple objectives. Third, diversity is typically treated as an auxiliary concern rather than a first-class optimization objective. While semantic deduplication and clustering-based filtering are commonly used as preprocessing steps \cite{abbas2023semdedup,xu2025quality,yu2023devil}, they are decoupled from quality and task relevance modeling, leading to suboptimal trade-offs between retaining diverse concepts and preserving task-critical samples.

This work focuses on optimizing the training subset selection strategy from a fixed large-scale multimodal dataset to enhance the overall zero-shot performance of multimodal pre-trained models such as CLIP across multiple downstream tasks. Unlike prior approaches that rely solely on general quality metrics or single-task feedback, we tackle a more practically significant and challenging problem: 

\textit{How to construct a task-aware data selection mechanism for multi-task multimodal pre-training that explicitly models sample-wise contributions to multiple tasks, while leveraging data quality, task relevance, and diversity as guidance to maximize overall multi-task performance? }

To address this, we propose TADS, a learnable framework that unifies data quality, task relevance, and diversity. Specifically: (1) Quality Assessment: We construct a comprehensive feature system integrating unimodal and cross-modal operators to resolve the limitations of fragmented quality metrics \cite{gadre2023datacomp}. (2) Task Relevance: We explicitly quantify sample utility by projecting candidate samples into a unified semantic space alongside task-specific prototypes. This yields interpretable relevance vectors that characterize the multi-task utility landscape, providing a dense guidance signal for the learnable selection policy. (3) Diversity Optimization: We introduce a cluster-based diversity factor that up-weights long-tail samples to prevent semantic bias. These dimensions are fused via a Data Value Network (DVN) to model non-linear interactions. Finally, we employ a bi-level feedback-driven optimization: the inner loop evaluates candidate subsets via a proxy model, while the outer loop updates the DVN using cluster-aware gradient estimation derived from multi-task performance feedback.
The main contributions are summarized as follows:
\begin{itemize}
    \item We propose a novel Task-Aware Data Selection Framework designed for multi-task multimodal pre-training. By taking the proposed task relevance vectors, diversity factors, and fine-grained quality scores as joint inputs, the framework employs a feedback-driven optimization mechanism to learn a data value function. This approach precisely estimates the value of each sample, thereby effectively aligning data selection with multiple downstream task objectives.
    \item We construct a systematic multimodal quality assessment system that integrates comprehensive unimodal and cross-modal operators. Through a hybrid learning strategy combining weak supervision and ground truth labels, the system produces reliable and fine-grained quality estimates for large-scale noisy multimodal data, serving as a task-agnostic quality prior within the proposed data selection framework.
    \item Experiments on CC12M demonstrate that, under a fixed pretraining budget, TADS achieves an average improvement of 1.0\% across multiple downstream tasks while using only 36\% of the training data. 
\end{itemize}

\section{Related Work}

\subsection{Task-Agnostic Multimodal Data Selection}
Early efforts primarily utilized heuristic filters, such as image resolution, caption length, and CLIP-based cosine similarity \cite{schuhmann2021laion, hessel2021clipscore, gadre2023datacomp}. While these methods mitigate coarse noise, they rely on handcrafted thresholds that struggle to capture non-linear quality dimensions \cite{mahmoud2024sieve}. Recent data-driven approaches, such as DFN \cite{fang2023data} and EcoDatum \cite{xu2025quality}, leverage supervised or probabilistic models to estimate a unified quality score. 

Although these methods excel at identifying globally well-formed samples, they treat selection as a task-independent preprocessing step. This leads to a centripetal bias towards generic vision-language manifolds, where samples are selected for their average aesthetic or alignment properties rather than their specific informative utility for heterogeneous downstream tasks.

\subsection{Task-Aware Data Selection}
To bridge the gap between pre-training and downstream application, recent works incorporate task-specific signals. Methods like FLYT \cite{shechter2025filter} and MGD \cite{engstrom2025optimizing} utilize performance feedback, while HYPE \cite{kim2024hype} and NormSim \cite{wang2024cliploss} quantify task relevance through semantic similarity to task datasets. 

These approaches are predominantly optimized for single-task scenarios. Their valuation functions are tightly coupled to a specific task distribution, lacking a mechanism for collaborative optimization across conflicting objectives. Consequently, a subset optimized for one task may inadvertently prune samples essential for the semantic coverage of another, leading to suboptimal multi-task generalization. Furthermore, aggressive task-targeting often sacrifices distributional diversity, a critical factor for model robustness \cite{liu2024tsds, yan2025coido}.

TADS transcends these boundaries by unifying intrinsic quality, multi-task relevance, and distributional diversity into a learnable policy. We employ a feedback-driven meta-learning mechanism to autonomously resolve the trade-offs between multiple task objectives, ensuring the selected subset is both high-utility and semantically diverse.

\begin{figure*}[t]
    \centering
    \includegraphics[width=1\linewidth]{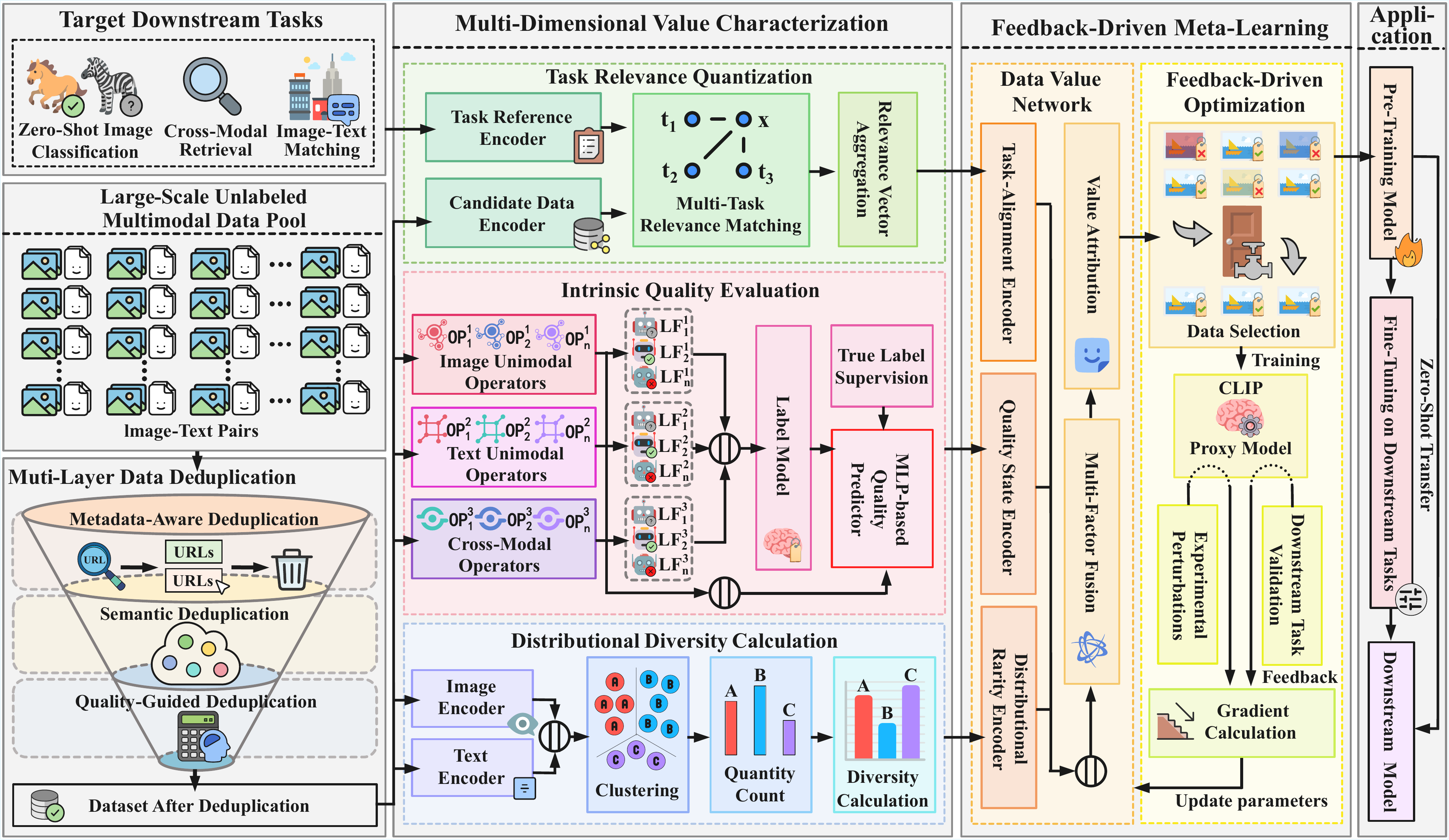}
    \caption{Overview of the proposed TADS framework. The framework operates through a cascade of three integrated stages: (1) \textbf{Multi-Layer Data Deduplication}, which physically filters redundancy; (2) \textbf{Multi-Dimensional Value Characterization}, which encodes data utility from quality, relevance, and diversity perspectives; (3) \textbf{Feedback-driven Meta-Learning}, a bi-level learning mechanism that autonomously updates the selection strategy based on proxy model feedback.}
    \label{fig2_introduction}
\end{figure*}

\section{Methodology}

As illustrated in Figure \ref{fig2_introduction}, TADS is designed to bridge the gap between large-scale, noisy multimodal data pools and specific downstream applications.

\subsection{Multi-Layer Data Deduplication}
\label{sec:deduplication}
Raw web-crawled data $\mathcal{U} = \{x_i\}_{i=1}^N$, where each sample $x_i = (I_i, T_i)$ consists of an image $I_i$ and its corresponding textual description $T_i$, undergoes a coarse-to-fine deduplication process. As shown in the data deduplication module of Figure \ref{fig2_introduction}, this process comprises three layers:

\textbf{Metadata-Aware Deduplication:} We first address physical redundancy by identifying exact duplicates via file hashes and source URLs. Instead of random retention, we maximize the information density of the unique set. Formally, let $\mathcal{G} = \{x_1, \dots, x_m\} \subset \mathcal{U}$ denote a collision set of samples sharing identical metadata identifiers. We retain the canonical instance $x^*$ that maximizes a heuristic initial quality score $S_{init}$:
\begin{equation}
\begin{aligned}
    x^* &= \operatorname*{arg\,max}_{x \in \mathcal{G}} S_{init}(x), \\
    S_{init}(x) &= \alpha_r \cdot \mathcal{R}(I) + \alpha_l \cdot \textit{Len}(T)
\end{aligned}
\end{equation}
where $\mathcal{R}(I)$ denotes the image resolution, $\textit{Len}(T)$ represents the text length, and $\alpha_r, \alpha_l$ are balancing coefficients.

\textbf{Semantic Deduplication:} To mitigate semantic redundancy, we map samples into a joint semantic space and apply a density-aware pruning strategy. 
Formally, we utilize the pre-trained CLIP encoder to extract image embeddings $z^I$ and text embeddings $z^T$, constructing a unified semantic vector $z^{joint} = [z^I; z^T]$.  We then employ the Mini-Batch K-Means algorithm to partition the dataset $\mathcal{U}$ into $M$ disjoint semantic clusters $\mathcal{C} = \{C_1, C_2, \dots, C_M\}$.

Within each cluster $C_j$, we rank samples based on the initial quality score $S_{init}$ and retain the top $K_j$ samples to construct the semantically deduplicated subset $S_{sem}$:
\begin{equation}
    S_{sem} = \bigcup_{j=1}^{M} \operatorname{Top}(C_j, S_{init}, K_j), K_j = \lceil \gamma \cdot |C_j| \rceil
\end{equation}
$|C_j|$ denotes the cardinality of the cluster, $\operatorname{Top}(\cdot, \cdot, K_j)$ represents the operation of selecting the $K_j$ highest-scoring samples, and $\gamma \in (0, 1]$ is the retention ratio to determine the subset size dynamically.

\textbf{Quality-Guided Deduplication:} The final stage eliminates fine-grained redundancies within local semantic clusters. To capture diverse forms of text duplication, we decompose redundancy into two distinct indicators: (1) Structural Redundancy ($\mathbb{I}_{struct}$), which captures near-identical character sequences using Edit Distance. (2) Semantic Redundancy ($\mathbb{I}_{sem}$), which identifies paraphrased content with high embedding similarity using the text encoder.
\begin{equation}
    \mathbb{I}_{struct}(x_i, x_j) = \mathbb{I}(\textit{EditDist}(T_i, T_j) < \tau_{edit})
\end{equation}
\begin{equation}
    \mathbb{I}_{sem}(x_i, x_j) = \mathbb{I}(\cos(z_i^T, z_j^T) > \tau_{sem})
\end{equation}
$\tau_{edit}$ represents the structural tolerance threshold, and $\tau_{sem}$ denotes the semantic equivalence threshold. $\mathbb{I}(\cdot)$ is the standard indicator function. We determine these hyperparameters via a pilot study on a held-out subset. By analyzing the trade-off curve between deduplication efficiency and semantic preservation, we identify the elbow point that maximizes redundancy removal while maintaining a false removal rate below 1\%.

We define the Composite Redundancy Indicator $\mathbb{I}_{red}(x_i, x_j)$. A pair is formally deemed redundant if it triggers either the structural or semantic condition:
\begin{equation}
    \mathbb{I}_{red}(x_i, x_j) = \mathbb{I}_{struct}(x_i, x_j) \lor \mathbb{I}_{sem}(x_i, x_j)
\end{equation}
For any identified redundant group $\mathcal{G}_{red} = \{x \mid \forall x_i, x_j \in \mathcal{G}_{red}, \mathbb{I}_{red}(x_i, x_j)=1\}$, we retain the canonical instance $x'$ that exhibits the highest visual grounding, quantified by the CLIP alignment score $S_{align}(x) = \cos(z^I, z^T)$:
\begin{equation}
    x' = \operatorname*{arg\,max}_{x \in \mathcal{G}_{red}} S_{align}(x)
\end{equation}
This operation yields the refined dataset $\mathcal{U}'$.

\subsection{Multi-Dimensional Value Characterization}
\label{sec:characterization}
We postulate that data utility is a vector composed of three orthogonal dimensions, each defining a distinct rejection boundary: Intrinsic Quality establishes the Signal-to-Noise baseline, filtering physically degraded or meaningless samples independent of task context. Task Relevance and Distributional Diversity jointly resolve the Utility-Redundancy dilemma within the semantic space. Specifically, Task Relevance acts as a centripetal force pulling selection towards task prototypes, while Distributional Diversity acts as a centrifugal force penalizing high-density regions. These dimensions are orthogonal: a sample can be high-quality yet irrelevant, or highly relevant yet redundant. TADS optimizes their intersection to select a subset that is simultaneously valid, useful, and informative.
\subsubsection{Intrinsic Quality Evaluation}
Quantifying the quality of uncurated web data is ill-posed due to the absence of ground truth and the multi-faceted nature of quality. We address this by constructing a high-dimensional feature system and employing a Hybrid Weak Supervision strategy.

\begin{figure}[t]
    \centering
    \includegraphics[width=1\linewidth]{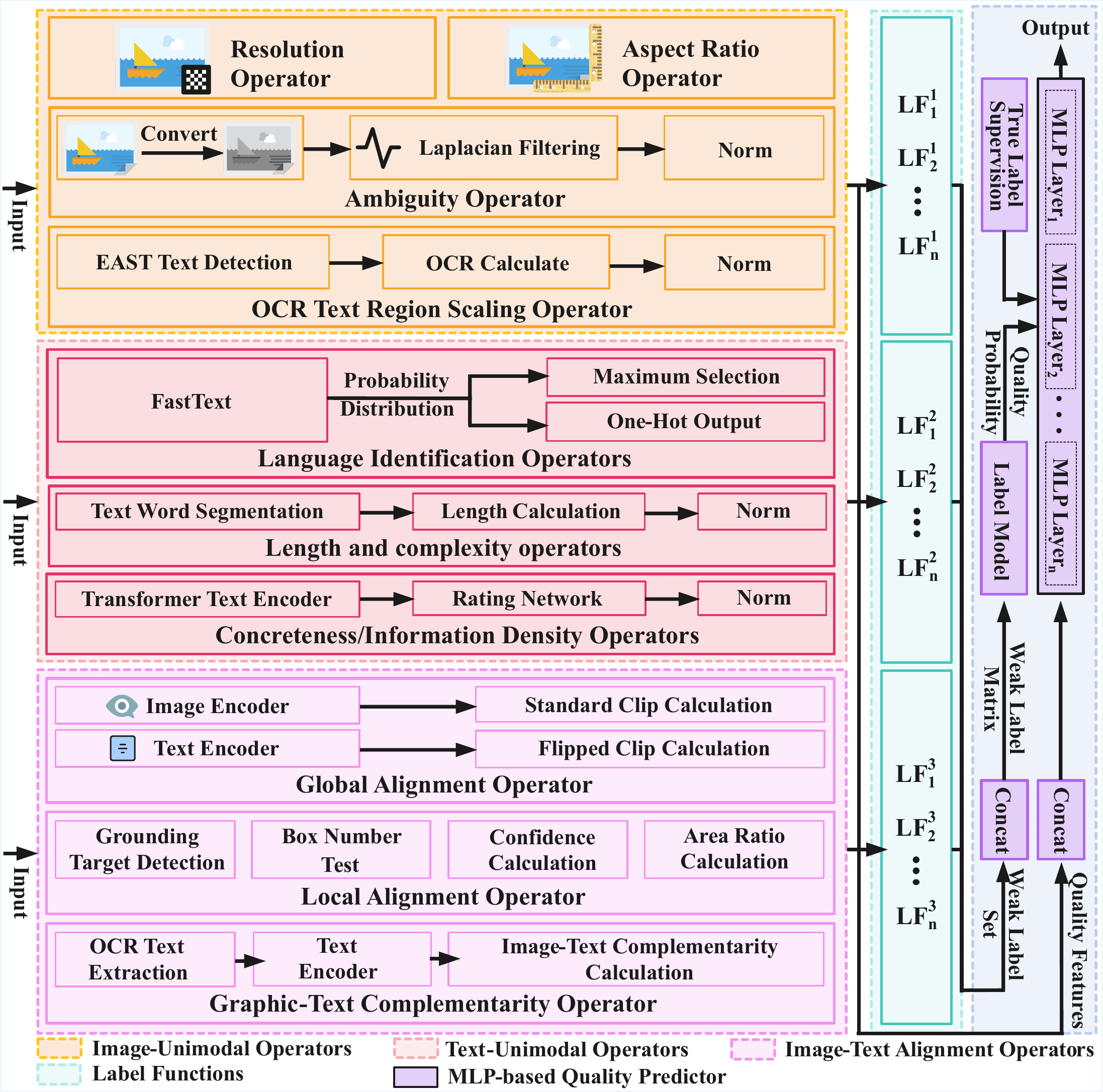}
    \caption{Architecture of the Intrinsic Quality Evaluator. It integrates comprehensive unimodal and cross-modal operators to encode quality features. The MLP predictor is optimized via a hybrid learning framework that synergizes Snorkel-based weak supervision with ground-truth guidance to quantify sample quality.}
    \label{fig3_introduction}
\end{figure}

\textbf{Fine-Grained Feature Extraction.} As illustrated in Figure \ref{fig3_introduction}, we move beyond simplistic heuristics by constructing a hierarchical operator system that maps raw inputs into a comprehensive feature vector $f(x) = [f_I(x); f_T(x); f_M(x)] \in \mathbb{R}^{d_f}$, where $d_f$ denotes the aggregate dimensionality of the fused operator features. This system operates across three distinct granularities: (1) \textbf{Image-Unimodal Operators ($f_I$):} We gauge visual integrity through a multi-stage pipeline. Besides standard resolution and aspect ratio metrics, we employ a Laplacian Filtering operator to quantify image ambiguity. To penalize low-utility visual clutter, we utilize the EAST detector to calculate the OCR Text Region Ratio, discarding samples dominated by dense, non-scene text. (2) \textbf{Text-Unimodal Operators ($f_T$):} We assess linguistic tractability and richness. The FastText classifier is deployed for robust language identification. To measure semantic capacity, we introduce a Concreteness/Information Density Operator, which utilizes a lightweight Transformer encoder followed by a rating network to distinguish substantive descriptions from vague or gibberish text. (3) \textbf{Cross-Modal Alignment Operators ($f_M$):} We disentangle alignment into global and local dimensions. We compute standard CLIP similarity alongside a Flipped-Image Consistency score to detect layout-sensitive noise and employ a Grounding Target Detection module to calculate the box number and confidence of text entities mapped to image regions. We explicitly model the information gain of the caption relative to the image's embedded OCR text, ensuring the text adds semantic value beyond reading text inside the image.

\textbf{Probabilistic Quality Inference.} Directly mapping $f(x)$ to a quality score is challenging without labels. We employ a hybrid approach combining Snorkel logic with perturbation-based supervision.
First, we define a set of heuristic labeling functions (LFs) based on statistical thresholds of $f(x)$. A Snorkel LabelModel estimates the latent accuracy of these LFs to generate probabilistic weak labels $\hat{y}^{weak}$.
Simultaneously, we construct a small True Label set $\mathcal{D}_{true}$ by sampling high-quality pairs and generating synthetic negatives via text shuffling and semantic mismatch injection.
Finally, to effectively distill the noisy weak signals while leveraging the precise ground truth, we train the MLP-based Quality Predictor $Q_\phi$ by minimizing a hybrid objective function $\mathcal{L}(\phi)$. This objective strictly enforces consistency with the weak probabilistic labels $\hat{y}^{weak}$ on the unlabeled pool $\mathcal{U}'$, while aligning with the gold-standard anchors in $\mathcal{D}_{true}$:
\begin{equation}
    \mathcal{L}_{sup} = \mathbb{E}_{(x,y) \sim \mathcal{D}_{true}} \left[ \mathcal{L}_{BCE}(q_\phi(x), y) \right]
\end{equation}
\begin{equation}
    \mathcal{L}_{weak} = \mathbb{E}_{x \sim \mathcal{U}'} \left[ \| q_\phi(x) - \hat{y}^{weak}(x) \|_2^2 \right]
\end{equation}
\begin{equation}
    \mathcal{L}(\phi) = \lambda_1 \cdot \mathcal{L}_{sup} + \lambda_2 \cdot \mathcal{L}_{weak}
\end{equation}
where $\mathcal{L}_{BCE}$ denotes the binary cross-entropy loss between the prediction $q_\phi(x)$ and the ground truth label $y$. The second term utilizes Mean Squared Error (MSE) to align the prediction with the soft targets $\hat{y}^{weak}(x)$ generated by the Label Model. Hyperparameters $\lambda_1$ and $\lambda_2$ balance the trade-off between supervision strength and weak label regularization. This optimization yields a robust, differentiable quality estimator $q_\phi(x) \in [0, 1]$ capable of generalizing beyond the heuristic rules.
\subsubsection{Task Relevance Quantization}
General quality does not imply utility for specific downstream tasks. To quantify task-specific value, we project candidate samples into a unified semantic space alongside task-specific semantic anchors.

Let $\mathcal{M} = \{M_1, \dots, M_K\}$ be the set of $K$ diverse downstream tasks. We represent each task $M_k$ by a constructed Task Support Set $\mathcal{S}_k$. This set consists of textual prototypes derived from the task's taxonomy. Formally, for a candidate sample $x$, we calculate its relevance score $r_k(x)$ by measuring its semantic proximity to the manifold defined by $\mathcal{S}_k$:
\begin{equation}
    r_k(x) = \frac{1}{|\mathcal{S}_k|} \sum_{p \in \mathcal{S}_k} \cos(E(x), E(p))
\end{equation}
where $E(\cdot)$ denotes the pre-trained multimodal encoder, and $p$ represents the textual prototype embeddings defining the task boundary.

Finally, to construct a globally comparable task-aware profile, we concatenate the scores from all tasks and apply $L_2$ normalization:
\begin{equation}
    h_r(x) = \frac{v}{\|v\|_2+ \epsilon}, v = [r_1(x), \dots, r_K(x)]^T
\end{equation}
$\epsilon$ is a tiny constant to prevent division by zero. The resulting Task Relevance Vector $h_r(x) \in \mathbb{R}^K$ explicitly encodes the sample's alignment with the multi-task objective landscape.

\subsubsection{Distributional Diversity Calculation}
To prevent the selection policy from overfitting to high-frequency semantic patterns and ignoring the long tail, we introduce a diversity mechanism grounded in the density of the joint embedding space.

We perform Mini-Batch K-Means clustering on the refined pool $\mathcal{U'}$ using the joint embeddings $z^{joint}$ to partition the data into semantic clusters. For each sample $x$, we identify its cluster $C_x$ and compute a diversity factor $d(x)$ based on the cluster cardinality $|C_x|$:
\begin{equation}
    d(x) = \left( \frac{1}{|C_x| + \epsilon} \right)^\delta
\end{equation}
where $\epsilon$ ensures numerical stability. The hyperparameter $\delta \ge 0$ serves as a density sensitivity coefficient that modulates the strength of distributional flattening. When $\delta = 0$, the factor collapses to $d(x)=1$, reducing the policy to a diversity-agnostic state that preserves the imbalanced data distribution. As $\delta$ increases, the penalty on high-density clusters becomes more severe. This forces the model to aggressively up-weight rare samples from sparse clusters, effectively shifting the training distribution from head-dominant to long-tail balanced.

\subsection{Feedback-Driven Meta-Learning}
\label{sec:optimization}
In TADS framework, the data selection policy is continuously refined through a closed-loop mechanism, guided by real-time performance feedback from a proxy learner to maximize generalization across downstream tasks.

\subsubsection{Data Value Network (DVN)}
The DVN functions as the parameterized policy $\pi_\theta$ that governs the selection distribution. It maps the multi-dimensional feature profile of a sample $x$ to a scalar utility score. As illustrated in the DVN module of Figure \ref{fig2_introduction}, we design a specialized multi-stream architecture to process heterogeneous inputs:

\textbf{Feature Encoding.} We employ distinct projection heads to map raw signals into a shared latent space $\mathbb{R}^d$: $q_\phi(x)$ and $d(x)$ are projected via learnable Feature Embedding Layers ($E_q, E_d$) to capture their non-linear contribution scales, while $h_r(x)$ is processed by a dedicated Task-Alignment Encoder ($E_r$) to extract inter-task dependencies.

\textbf{Multi-Factor Fusion.} The encoded representations are concatenated and fed into a Fusion MLP to model higher-order interactions. To ensure the output represents a valid probabilistic selection signal, we apply a Sigmoid activation:
\begin{equation}
    v_\theta(x) = \sigma \left( \text{MLP}_{fuse} \left( [E_q(q_\phi), E_r(h_r), E_d(d)] \right) \right)
\end{equation}
This score $v_\theta(x) \in (0, 1)$ explicitly quantifies the selection probability of each sample. It governs the construction of the training subset $S(\theta)$ by treating the inclusion of each sample as a Bernoulli trial.

\subsubsection{The Feedback-Driven Optimization}
\label{sec:optimization}
Since data value lacks explicit supervision, we rely on the downstream performance as a delayed reward signal. We establish a feedback mechanism comprising three sequential phases: \textit{Simulation}, \textit{Evaluation}, and \textit{Adaptation}.

\textbf{Proxy-Based Performance Simulation.} To ensure the learned data selection policy generalizes effectively to the final pre-training stage, we employ a performance-aligned Proxy Model $\omega$ as a simulator. We adopt the ViT-B/16 architecture to maintain high fidelity between the proxy's feedback and the target model’s actual behavior, thereby ensuring accurate data valuation. In each optimization step, we sample a subset $S(\theta)$ based on the current policy $v_\theta$. The proxy model is then rapidly adapted on $S(\theta)$ to minimize the contrastive loss:
\begin{equation}
    \omega^*(\theta) = \arg\min_{\omega} \mathbb{E}_{x \in S(\theta)} [\mathcal{L}_{InfoNCE}(x; \omega)]
\end{equation}
$\omega^*$ is the adapted proxy, which acts as a probe to measure data utility. This step simulates the learning trajectory of a model trained on the currently selected data distribution.

\textbf{Feedback Acquisition via Multi-Task Evaluation.} To guide the optimization, we define a Meta-Reward function $\mathcal{J}(\cdot)$ over the global score vector $V = \{v_\theta(x) \mid x \in \mathcal{U}'\}$. We evaluate the adapted proxy model $\omega^*$ on the task validation sets and aggregate the metrics:
\begin{equation}
    \mathcal{J}(V) = \sum_{k=1}^K \varphi_k \cdot \textit{Eval}_k(\omega^*(V); \mathcal{S}_k)
\end{equation}
where $\mathcal{S}_k$ denotes the task support set, and $\varphi_k$ balances multi-task objectives. While the ultimate goal is to optimize parameters $\theta$, $\mathcal{J}$ is directly a function of the intermediate selection scores $V$.

\textbf{Cluster-Aware Gradient Estimation \& Adaptation.}  Since the selection process involves discrete sampling, the objective is non-differentiable. To propagate the feedback signal back to $\theta$, we treat the selection of each sample as a stochastic action and employ a Cluster-Aware Gradient Estimation strategy.

We model the selection action for sample $x$ as a random variable $m_x \sim \text{Bernoulli}(v_\theta(x))$. To estimate the policy gradient, we leverage the semantic clusters $\{C_i\}$ as control variates. We apply a perturbation vector $\delta_{C_i}$ to the scores of samples in cluster $C_i$ and measure the sensitivity of the global reward. The approximate gradient of the reward with respect to the cluster scores is estimated as:
\begin{equation}
    g_{C_i} \approx \frac{\mathcal{J}(V + \delta_{C_i}) - \mathcal{J}(V)}{\sigma}
\end{equation}
Here, $g_{C_i}$ represents the \textit{advantage signal} for cluster $C_i$. Finally, using the Chain Rule, we update the DVN parameters $\theta$ via a REINFORCE-based policy gradient objective:
\begin{equation}
    \mathcal{L}_{policy}(\theta) = - \sum_{C_i} g_{C_i} \sum_{x \in C_i} \log v_\theta(x) + \beta \|\theta\|_2^2
\end{equation}
By minimizing $\mathcal{L}_{policy}$, the network increases the log-likelihood of selecting samples in clusters with positive contribution and suppresses those with negative impact, effectively aligning the selection distribution with multi-task generalization.

\section{Experiments}
\label{sec:experiments}

\subsection{Experimental Setup}
\label{sec:setup}

\textbf{Datasets and Metrics.}
We utilize \textbf{CC12M} \cite{changpinyo2021conceptual}, a large-scale noisy dataset containing approximately 11 million image-text pairs, as the uncurated candidate pool $\mathcal{U}$.
To evaluate multi-task generalization, we employ a diverse suite of downstream benchmarks:
(1) \textbf{Classification:} ImageNet-1K \cite{deng2009imagenet} and CIFAR-100 \cite{krizhevsky2009learning}, reporting Zero-shot Top-1 (Top-1) and Top-5 (Top-5) Accuracy.
(2) \textbf{Cross-Modal Retrieval:} MS-COCO \cite{lin2014microsoft} and Flickr30k \cite{plummer2015flickr30k}, reporting Text Retrieval Recall@1 (TR@1) and Image Retrieval Recall@1 (IR@1).

\textbf{Baselines.}
We benchmark TADS against a comprehensive set of 10 state-of-the-art selection strategies, categorized into two paradigms to highlight methodological differences:
(1) \textbf{Task-Agnostic Multimodal Data Selection:} Methods that filter data based on statistical rules or redundancy, including Text Complexity \cite{gadre2023datacomp}, SemDeDup \cite{abbas2023semdedup}, CLIP-Score \cite{hessel2021clipscore}, T-MARS \cite{maini2023t}, SIEVE \cite{mahmoud2024sieve}, s-CLIPLoss \cite{wang2024cliploss}, and EcoDatum \cite{xu2025quality}.
(2) \textbf{Task-Aware Multimodal Data Selection:} Advanced methods that leverage task signals, including HYPE \cite{kim2024hype} (specificity-based) and FLYT+SCS \cite{shechter2025filter} (soft sampling).
Additionally, we report \textit{No Filtering} (using the full 10.97M pool) as the performance baseline.

\begin{table*}[t]
\centering
\caption{\textbf{Zero-shot performance comparison on downstream tasks.} All models are pre-trained on CC12M with a fixed computational budget of 128M samples. \textbf{Bold} and \underline{underlined} values indicate the \textbf{best} and \underline{second-best} results, respectively.}
\label{tab:main_results}
\resizebox{\textwidth}{!}{
\begin{tabular}{l|c|c|cc|cc|cc|cc|c}
\toprule
\multirow{2}{*}{\textbf{Method}} & \multirow{2}{*}{\textbf{Type}} & \multirow{2}{*}{\textbf{Data Size}} & \multicolumn{2}{c|}{\textbf{ImageNet-1K}} & \multicolumn{2}{c|}{\textbf{CIFAR-100}} & \multicolumn{2}{c|}{\textbf{MS-COCO}} & \multicolumn{2}{c|}{\textbf{Flickr30K}} & \multirow{2}{*}{\textbf{AVG.}} \\
\cmidrule(lr){4-5} \cmidrule(lr){6-7} \cmidrule(lr){8-9} \cmidrule(lr){10-11}
 & & & Top-1 & Top-5 & Top-1 & Top-5 & TR@1 & IR@1 & TR@1 & IR@1 & \\
\midrule
\textit{No Filtering (Baseline)} & - & $\sim$10.97M & 28.2 & 53.1 & 25.4 & 58.1 & 27.3 & 18.3 & 46.5 & 35.2 & 36.5 \\
\midrule
Text Complexity \cite{gadre2023datacomp} & Task-Agnostic & $\sim$8.56M & 28.9 & 54.3 & 26.0 & 58.8 & 27.4 & 18.8 & 47.4 & 35.8 & 37.2 \\
SemDeDup \cite{abbas2023semdedup} & Task-Agnostic & $\sim$4.39M & 29.6 & 54.9 & 26.5 & 59.2 & 28.9 & 19.2 & 48.1 & 36.1 & 37.8 \\
CLIP-Score \cite{hessel2021clipscore} & Task-Agnostic & $\sim$6.91M & 30.1 & 55.3 & 27.2 & 60.5 & 30.7 & 20.6 & 51.9 & 38.8 & 39.4 \\
T-MARS \cite{maini2023t} & Task-Agnostic & $\sim$5.49M & 30.8 & 56.4 & 27.8 & 61.0 & 30.2 & 20.2 & 50.8 & 38.3 & 39.4 \\
SIEVE \cite{mahmoud2024sieve} & Task-Agnostic & $\sim$3.29M & 31.7 & 57.0 & 28.5 & 62.5 & 26.6 & 19.0 & 45.2 & 36.7 & 38.4 \\
s-CLIPLoss \cite{wang2024cliploss} & Task-Agnostic & $\sim$6.58M & 32.3 & 58.5 & 29.7 & 64.1 & 32.4 & 21.8 & 54.7 & 40.5 & 41.8 \\
EcoDatum \cite{xu2025quality} & Task-Agnostic & $\sim$4.39M & 36.2 & 62.2 & 34.0 & 69.3 & 35.5 & 24.1 & 58.4 & 43.1 & 45.4 \\
HYPE \cite{kim2024hype} & Task-Aware & $\sim$3.29M & 36.5 & 62.1 & 32.5 & 67.4 & 32.1 & 22.0 & 53.2 & 40.1 & 43.2 \\
HYPE + s-CLIPLoss \cite{kim2024hype} & Task-Aware & $\sim$2.52M & 38.2 & 63.8 & 33.8 & 68.9 & 34.2 & 23.1 & 56.5 & 42.0 & 45.1 \\
FLYT + SCS \cite{shechter2025filter} & Task-Aware & $\sim$10.97M & \underline{39.5} & \textbf{66.5} & \underline{36.8} & \textbf{72.6} & \underline{36.9} & \underline{25.2} & \underline{59.8} & \underline{45.5} & \underline{47.9} \\
\midrule
\textbf{TADS (Ours)} & \textbf{Task-Aware} & \textbf{$\sim$3.95M} & \textbf{40.7} & \underline{66.1} & \textbf{38.6} & \underline{72.1} & \textbf{38.1} & \textbf{26.8} & \textbf{60.9} & \textbf{47.5} & \textbf{48.9} \\
\bottomrule
\end{tabular}
}
\end{table*}
\subsection{Main Results: Zero-Shot Transfer Performance}
As shown in Table \ref{tab:main_results}, we analyze the results from three perspectives: data efficiency, the precision-coverage trade-off, and semantic alignment capabilities.

\textbf{Breaking the Agnostic Ceiling.} Task-agnostic methods improve over the unfiltered baseline by removing noise but hit a clear performance plateau. For instance, EcoDatum peaks at 36.2\% on ImageNet Top-1. This limitation stems from their objective of identifying universally high-quality samples, which often retains visually pristine but task-irrelevant data. By introducing explicit Task Relevance, TADS breaks this ceiling, achieving a substantial +4.5\% gain over the best task-agnostic method on ImageNet Top-1.

\textbf{Efficiency and The Precision-Coverage Trade-off.} A critical comparison with the strongest baseline, FLYT+SCS, highlights the efficiency of our approach. TADS achieves SOTA Top-1 performance using only $\sim$3.95M samples, whereas FLYT requires the full $\sim$10.97M dataset. While TADS dominates in Top-1 Accuracy—outperforming FLYT by +1.2\% on ImageNet and +1.8\% on CIFAR-100—it marginally lags behind on Top-5 Accuracy. This slight drop of 0.4\% is expected: FLYT's soft sampling of the full dataset provides broader coverage, whereas TADS's aggressive selection sharpens the decision boundary for the most probable class. Given that Top-1 is the harder and more discriminative metric, this trade-off validates TADS's superior feature extraction capability under strict budget constraints.

\textbf{Semantic Completeness in Retrieval.} TADS demonstrates robust cross-modal alignment. On Flickr30K TR@1, TADS achieves 60.9\%, surpassing the best baseline by +1.1\%. Unlike specificity-focused methods, which prioritize object-centric images and consequently suffer in retrieval tasks, TADS incorporates a Diversity Factor. This ensures the retention of semantically rich samples necessary for capturing complex image-text relationships, achieving the highest average performance (48.9\%) across all metrics.
\begin{table*}[t]
\centering
\small 
\caption{\textbf{Component-wise ablation study of TADS.} We incrementally add Intrinsic Quality ($Q$), Task Relevance ($R$), Distributional Diversity ($D$), and FDO to the baseline. All models are pre-trained on CC12M with a fixed budget of 128M samples seen.}
\label{tab:ablation_components}
\resizebox{0.95\textwidth}{!}{ 
\begin{tabular}{l|cccc|c|cc|cc|cc|cc}
\toprule
\multirow{2}{*}{\textbf{Method Variant}} & \multicolumn{4}{c|}{\textbf{Components}} & \multirow{2}{*}{\textbf{\shortstack{Data\\Retained}}} & \multicolumn{2}{c|}{\textbf{ImageNet-1K}} & \multicolumn{2}{c|}{\textbf{CIFAR-100}} & \multicolumn{2}{c|}{\textbf{MS-COCO}} & \multicolumn{2}{c}{\textbf{Flickr30K}} \\
 & $Q$ & $R$ & $D$ & FDO & & Top-1 & Top-5 & Top-1 & Top-5 & TR@1 & IR@1 & TR@1 & IR@1 \\
\midrule
\textit{No Filtering (Baseline)} & - & - & - & - & 100\% & 28.2 & 53.1 & 25.4 & 58.1 & 27.3 & 18.3 & 46.5 & 35.2 \\
\midrule
Quality Only & \checkmark & - & - & - & $\sim$47\% & 32.5 & 58.1 & 30.2 & 65.0 & 31.5 & 20.2 & 54.5 & 39.8 \\
+ Task Relevance & \checkmark & \checkmark & - & - & $\sim$28\% & 35.1 & 60.9 & 33.5 & 68.8 & 32.2 & 21.8 & 55.0 & 40.5 \\
+ Diversity & \checkmark & \checkmark & \checkmark & - & $\sim$31\% & 35.6 & 61.7 & 33.8 & 69.2 & 33.8 & 23.0 & 57.2 & 42.2 \\
\textbf{TADS (Full)} & \checkmark & \checkmark & \checkmark & \checkmark & $\sim$36\% & \textbf{40.7} & \textbf{66.1} & \textbf{38.6} & \textbf{72.1} & \textbf{38.1} & \textbf{26.8} & \textbf{60.9} & \textbf{47.5} \\
\bottomrule
\end{tabular}
}
\end{table*}

\begin{figure}[t]
    \centering
    \includegraphics[width=0.95\linewidth]{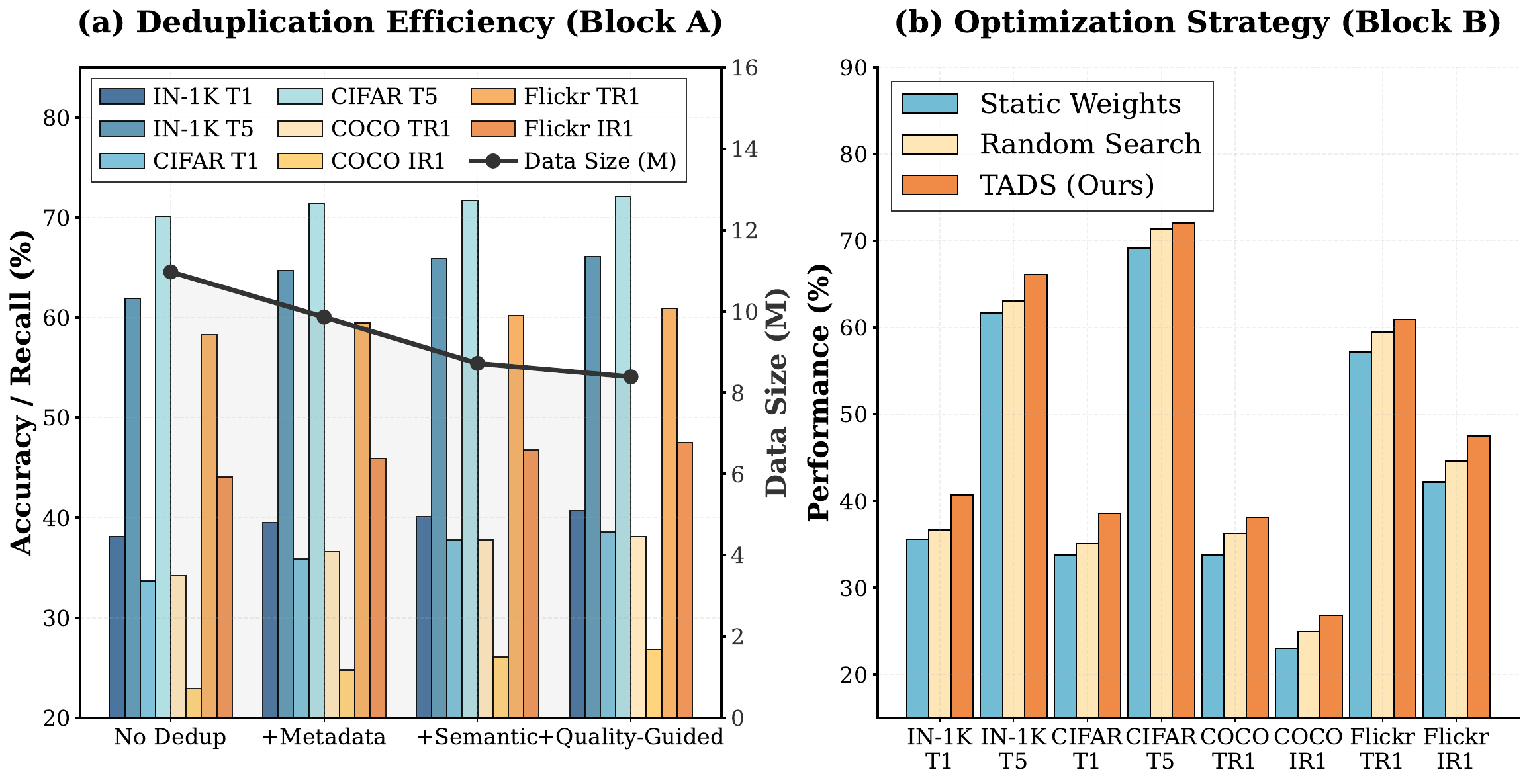}
    \caption{Detailed analysis of deduplication and optimization. (a) Deduplication Efficiency: Performance consistently improves as redundancy is removed, even as the data size decreases significantly. (b) Optimization Strategy: TADS outperforms heuristic and random search methods across all metrics, validating the effectiveness of the gradient estimation.}
    \label{fig4_introduction}
\end{figure}
\subsection{Ablation Study}
We conduct comprehensive ablation studies to analyze the contribution of each component in TADS and justify our design choices regarding deduplication and optimization.

\textbf{Effect of Data Valuation Components.} Table \ref{tab:ablation_components} presents a component-wise ablation by progressively introducing Intrinsic Quality ($Q$), Task Relevance ($R$), Distributional Diversity ($D$), and FDO. Starting from the unfiltered baseline, introducing Intrinsic Quality yields a substantial improvement, confirming the necessity of removing low-quality and noisy web-harvested samples.

Adding Task Relevance leads to further gains while significantly reducing the data scale to $\sim 28\%$. This validates that generic high-quality data is not equivalent to high-utility data, and task-aware selection can achieve superior efficiency. Incorporating Distributional Diversity increases the retention ratio to $\sim$31\% and brings additional improvements, indicating that preserving the semantic long-tail is crucial for complex cross-modal understanding.

Notably, enabling FDO provides the most significant boost, yielding a peak Top-1 accuracy of $40.7\%$ on ImageNet-1K. In this stage, the data retention ratio is adaptively adjusted to $\sim$36\%. This demonstrates that the DVN does not rely on rigid manual thresholds; instead, it autonomously learns to expand or contract the training subset to optimize the dynamic balance between data scale and multi-task utility.

\textbf{Effect of Deduplication Strategy.} Figure \ref{fig4_introduction} (Block A) visualizes the trajectory of performance gains through our multi-layer deduplication pipeline. We observe a consistent upward trend in downstream accuracy as redundancy is progressively removed. Crucially, as indicated by the descending black line, while Metadata and Semantic deduplication effectively prune the dataset volume, the final Quality-Guided Deduplication step provides the most critical boost. As shown in the figure, this strategy achieves the optimal trade-off, peaking at 40.7\% ImageNet Top-1, confirming that refining the information density of the candidate pool is a prerequisite for effective high-level selection.

\textbf{Comparison of Optimization Strategies.} Figure \ref{fig4_introduction} (Block B) illustrates the performance gap between different policy learning methods. The visual comparison clearly demonstrates the limitations of Static Weights and Random Search, which struggle to find the optimal policy in a high-dimensional search space. In contrast, our Cluster-Aware Gradient Estimation consistently surpasses these baselines across all metrics. This empirically validates that our gradient approximation effectively leverages proxy feedback to navigate the non-differentiable selection landscape, aligning the data policy with multi-task objectives.

\section{Conclusion}
In this work, we present TADS framework that advances multi-task multimodal pre-training by unifying intrinsic quality, task relevance, and distributional diversity in an adaptive, feedback-optimized manner. Our experiments show that, under a fixed pretraining budget, TADS achieves an average improvement of 1.0\% across multiple downstream tasks while using only 36\% of the training data, demonstrating superior zero-shot generalization compared to full-dataset training. This establishes TADS as a foundation for efficient and robust multimodal learning. Future work will focus on developing task relevance estimation robust to spurious correlations in sparse-definition regimes, improving error-tolerant deduplication and diversity modeling, and enabling efficient adaptation to dynamic downstream task sets.


\section*{Impact Statement}
This paper presents work whose goal is to advance the field of Machine Learning. There are many potential societal consequences of our work, none which we feel must be specifically highlighted here.

\nocite{langley00}

\bibliography{example_paper}
\bibliographystyle{icml2026}

\newpage
\appendix
\onecolumn
\section{Formal Algorithm for FDO}
To precisely characterize the technical implementation of our proposed selection mechanism, we formally define the FDO procedure in this section.
\begin{algorithm}[ht]
\caption{FDO for Data Selection}
\label{alg:tads_core}
\begin{algorithmic}[1]
\REQUIRE 
    Refined candidate pool $\mathcal{U}' = \{x_i\}_{i=1}^N$; 
    Target downstream tasks $\mathcal{T} = \{T_1, \dots, T_K\}$; 
    Feature extractor $\Phi$ (extracts quality $q$, relevance $h$, diversity $d$);
    Initial DVN parameters $\theta$; Proxy model $\mathcal{M}_\omega$.
\ENSURE Optimized DVN parameters $\theta^*$ for final data selection.

\STATE \textbf{Initialize:} Feature states $S = \{s_i = [q_i, h_i, d_i] \mid x_i \in \mathcal{U}'\}$ via $\Phi$.

\WHILE{not converged}
    \STATE \textit{// 1. Selection Strategy (Bernoulli Sampling)}
    \STATE Compute value scores $V = \{v_\theta(s_i)\}_{i=1}^N$ using the current DVN,
    \STATE where $v_\theta(s_i) \in (0,1)$ denotes the selection probability.
    \FOR{each sample $x_i \in \mathcal{U}'$}
        \STATE Draw a random number $r_i \sim \text{Uniform}(0,1)$.
        \STATE Set selection mask 
        \[
            m_i =
            \begin{cases}
                1, & \text{if } r_i \le v_\theta(s_i), \\
                0, & \text{otherwise}.
            \end{cases}
        \]
    \ENDFOR
    \STATE Define training subset $\mathcal{S}_\theta = \{x_i \in \mathcal{U}' \mid m_i = 1\}$.

    \STATE \textit{// 2. Inner Loop: Proxy Model Simulation}
    \STATE Reset proxy model parameters $\omega \leftarrow \omega_0$.
    \STATE Update $\omega$ by minimizing contrastive loss on $\mathcal{S}_\theta$:
    \STATE $\omega^*(\theta) = \arg\min_{\omega} \mathcal{L}_{pretrain}(\mathcal{M}_\omega; \mathcal{S}_\theta)$.

    \STATE \textit{// 3. Outer Loop: Meta-Feedback and Reward}
    \STATE Evaluate $\mathcal{M}_{\omega^*(V)}$ on validation sets of multiple tasks $\mathcal{T}$, where $V = \{v_\theta(x) \mid x \in \mathcal{U}'\}$
    \STATE Compute Meta-Reward: $\mathcal{J}(V) = \sum_{k=1}^K \varphi_k \cdot \textit{Eval}_k(M_{\omega^*(V)}; T_k)$. 
    
    \STATE \textit{// 4. Cluster-Aware Gradient Estimation}
    \FOR{each cluster $C_i$}
        \STATE Apply perturbation $\delta_{C_i}$ of magnitude $\sigma$ to value scores in $C_i$.
        \STATE Measure feedback shift: $\Delta \mathcal{J}_{C_i} = \mathcal{J}(V + \delta_{C_i}) - \mathcal{J}(V)$
        \STATE Compute approximate cluster gradient: $g_{C_i} = \Delta \mathcal{J}_{C_i} / \sigma$
    \ENDFOR
    \STATE Aggregate cluster gradients with REINFORCE-like weighting and regularization:
    \STATE $\mathcal{L}_{policy}(\theta) = - \sum_{C_i} g_{C_i} \sum_{x \in C_i} \log v_\theta(x) + \beta \|\theta\|_2^2$
    
    \STATE \textit{// 5. DVN Parameter Update}
    \STATE Update DVN parameters: $\theta \leftarrow \theta - \eta \nabla_\theta \mathcal{L}_{policy}(\theta)$

\ENDWHILE

\STATE \textbf{Final Selection:} $\mathcal{S}^* = \{x_i \in \mathcal{U}' \mid v_{\theta^*}(s_i) > \tau \}$.
\end{algorithmic}
\end{algorithm}

\section{Computational Efficiency and Scaling Analysis}
To further demonstrate the computational efficiency of our proposed \textbf{TADS} (Task-Aware Data Selection) framework, we conduct a scaling analysis by examining model performance under varying fixed training budgets, measured in terms of the total number of training samples seen.
\subsection{Performance under Fixed Computational Budgets}
\begin{figure*}[ht]
    \centering
    \includegraphics[width=1\linewidth]{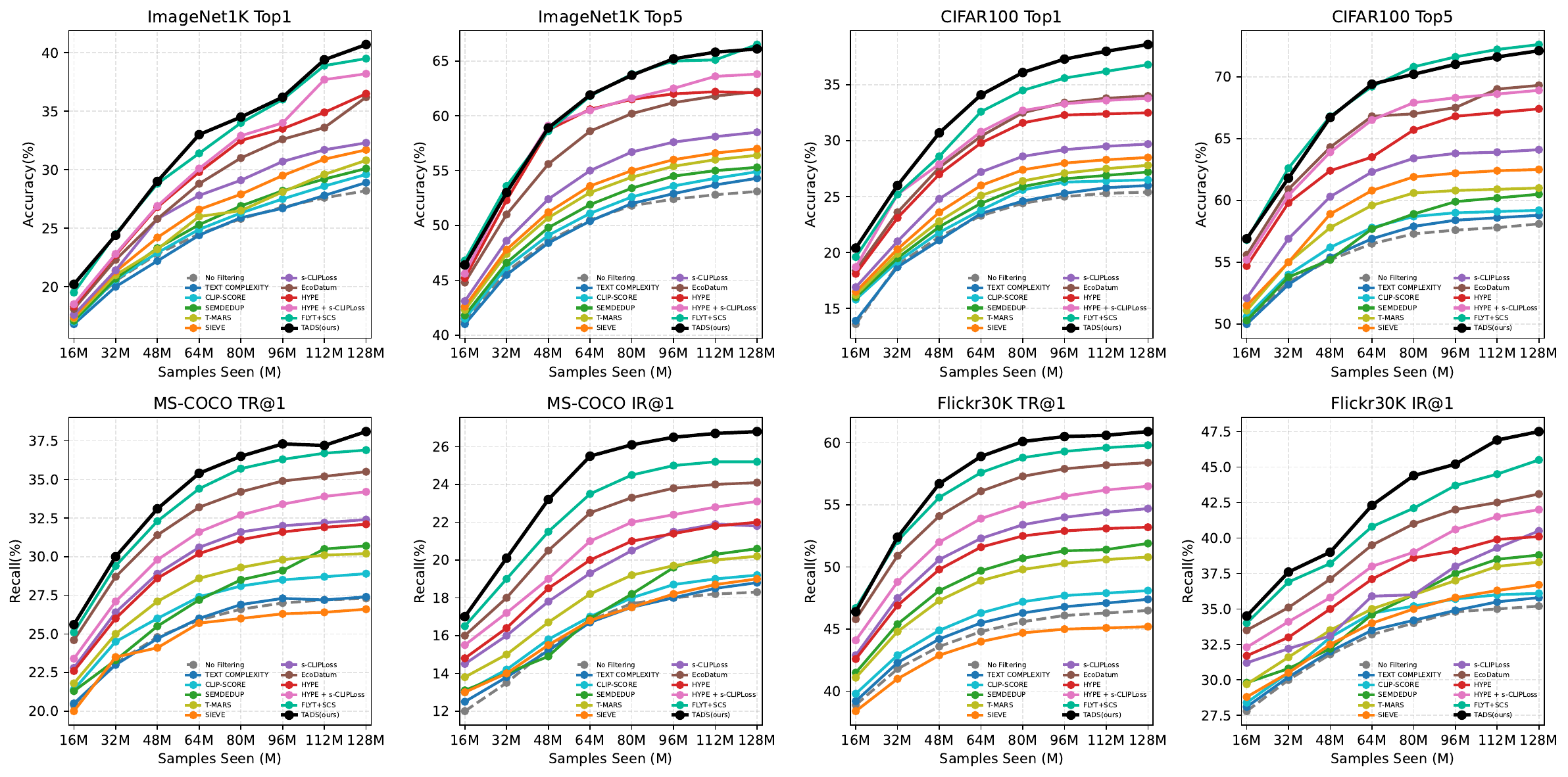}
    \caption{Zero-shot performance comparison on 4 downstream tasks across different training scales (from 16M to 128M samples seen). TADS consistently achieves higher accuracy and recall with the same computational budget compared to state-of-the-art data selection methods.}
    \label{fig5_appendix}
\end{figure*}

In large-scale multimodal pre-training, the number of training samples seen serves as a widely adopted proxy for computational consumption. We compare TADS against 11 representative baseline methods, covering a broad spectrum of data selection strategies, including heuristic-based approaches (e.g., CLIP-Score, Text Complexity), data-driven but task-agnostic methods (e.g., EcoDatum), and recent task-aware techniques (e.g., FLYT, HYPE). All methods are evaluated under fixed budgets ranging from 16M to 128M samples seen.
As illustrated in Figure \ref{fig5_appendix}, TADS consistently exhibits significant advantages:
\begin{itemize}
    \item \textbf{Vertical Advantage (Higher Accuracy):} At each fixed budget level (16M, 32M, 48M, \ldots, 128M), TADS achieves the best or near-best performance across all eight evaluation metrics. For example, on ImageNet-1K Top-1 accuracy and MS-COCO TR@1 recall, TADS maintains a clear margin over the strongest competing methods (e.g., Ecodatum or FLYT+SCS) throughout training.
    
    \item \textbf{Horizontal Advantage (Lower Cost):} To reach a given target performance, TADS requires substantially fewer training samples. In particular, TADS trained with approximately 48M samples matches or exceeds the performance of the \emph{No Filtering} baseline and standard \emph{CLIP-Score} filtering trained with the full 128M samples. This corresponds to an approximate \textbf{2.6$\times$} improvement in data efficiency under the same computational budget definition.
\end{itemize}

\subsection{Multi-Task Generalization Efficiency}
The efficiency of TADS is further reflected in its ability to generalize across multiple heterogeneous tasks. While several baseline methods achieve competitive performance on individual tasks (e.g., ImageNet classification), their performance under fixed computational budgets often fluctuates or saturates on retrieval benchmarks such as Flickr30K and MS-COCO.

In contrast, TADS leverages its \textit{Task Relevance Quantization} and \textit{Feedback-driven Meta-learning} mechanisms to prioritize samples that provide joint utility across the entire task suite. On the CC12M dataset, TADS achieves state-of-the-art zero-shot performance while utilizing only \textbf{36\%} of the original training data, outperforming all competing methods by an average margin of 1.0\% across evaluated tasks. These results indicate that task-aware data selection offers a more resource-efficient alternative to generic quality-based filtering when training versatile multimodal foundation models.

\section{Derivation of Cluster-Aware Feedback Optimization}
In this section, we provide the formal derivation for the FDO mechanism used in TADS. We detail how the discrete selection problem is transformed into a differentiable learning objective via cluster-based approximation.

\subsection{Problem Formulation}
Let the data selection policy be defined by the DVN with parameters $\theta$. For each sample $x$ in the candidate pool $\mathcal{U}$, the DVN outputs a selection probability $v_\theta(x) \in (0,1)$. The selection action is a discrete random variable sampled from a Bernoulli distribution:$$m_x \sim \text{Bernoulli}(v_\theta(x))$$
The objective is to maximize the expected multi-task reward $\mathcal{J}$ over the distribution of selected subsets:$$\max_\theta \mathbb{E}_{m \sim \pi_\theta} [\mathcal{J}(\mathcal{S}_m)]$$
where $\pi_\theta$ denotes the joint probability distribution of the selection mask $m$, and $\mathcal{S}_m = \{x \mid m_x = 1\}$ is the selected subset.

\subsection{The Non-Differentiability Challenge}
Directly optimizing this objective is challenging because the reward signal $\mathcal{J}$ (downstream performance) is a non-differentiable black-box function of the discrete mask $m$. Standard backpropagation cannot flow through the sampling step $m_x \sim \text{Bernoulli}(v_\theta(x))$.

While the REINFORCE algorithm allows for gradient estimation via $\nabla_\theta \mathcal{J} = \mathbb{E}[R(m) \cdot \nabla_\theta \log \pi_\theta(m)]$, applying it at the sample level is computationally prohibitive because we only receive a single scalar reward $R(m)$ for the entire subset, leading to extreme variance and the credit assignment problem.

\subsection{Cluster-Level Advantage Estimation}
To resolve the credit assignment problem and reduce variance, we assume that samples within the same semantic cluster $C_i$ share similar utility characteristics. We aim to estimate the sensitivity of the global reward $\mathcal{J}$ with respect to the collective score of a cluster, denoted as $V_{C_i}$.We treat the reward function $\mathcal{J}$ as a function of the continuous score vector $V = \{v_\theta(x)\}$. Using a first-order Taylor expansion, a perturbation $\delta_{C_i}$ applied to the scores of cluster $C_i$ results in:$$\mathcal{J}(V + \delta_{C_i}) \approx \mathcal{J}(V) + \nabla_{V_{C_i}} \mathcal{J}(V)^T \cdot \delta_{C_i}$$where $\nabla_{V_{C_i}} \mathcal{J}(V)$ represents the gradient of the reward with respect to the cluster scores. Assuming a uniform perturbation magnitude $\sigma$ along the cluster direction (i.e., $\delta_{C_i} = \vec{\sigma}$), we can rearrange this to approximate the gradient (advantage signal):$$\nabla_{V_{C_i}} \mathcal{J}(V) \approx \frac{\mathcal{J}(V + \delta_{C_i}) - \mathcal{J}(V)}{\sigma}$$This matches Equation 16 in the main text, where $g_{C_i}$ serves as the estimated gradient (or advantage) for cluster $C_i$:$$g_{C_i} \approx \frac{\Delta \mathcal{J}_{C_i}}{\sigma}$$

\subsection{Derivation of the Cluster-Aware Policy Gradient}
To formulate the parameter update, we apply the Policy Gradient Theorem. The derivation proceeds in three steps to bridge the gap between the standard REINFORCE algorithm and our cluster-aware approximation. 
\begin{enumerate}
    \item \textbf{The Log-Derivative Trick:} First, we apply the standard log-derivative trick to express the gradient of the expected reward:$$\nabla_\theta \mathcal{J} = \mathbb{E}_{m} [ R(m) \cdot \nabla_\theta \log \pi_\theta(m) ]$$
    \item \textbf{Factorization of the Policy:} Since the selection of each sample is an independent Bernoulli trial, the joint probability $\pi_\theta(m)$ factorizes. Consequently, the gradient of the log-probability decomposes into a sum over individual samples:$$\nabla_\theta \log \pi_\theta(m) = \sum_{x \in \mathcal{U}} \nabla_\theta \log P(m_x | \theta)$$Substituting this back, the standard gradient is $\mathbb{E}_{m} [ R(m) \cdot \sum_{x} \nabla_\theta \log v_\theta(x) ]$.
    
    \item \textbf{Cluster-Based Approximation:} The standard estimator uses the noisy global reward $R(m)$ for all samples. We introduce a Cluster Assumption: the contribution of samples in cluster $C_i$ is better approximated by the specific cluster advantage $g_{C_i}$ derived in Sec C.2.Replacing the global reward with the local advantage for each cluster group, we obtain the approximate gradient:$$\nabla_\theta \mathcal{J} \approx \sum_{C_i} g_{C_i} \sum_{x \in C_i} \nabla_\theta \log v_\theta(x)$$
\end{enumerate}

\subsection{Policy Loss Formulation}In standard deep learning frameworks, parameters are optimized by minimizing a loss function via gradient descent, rather than maximizing an objective via gradient ascent. To align with this paradigm, we define the policy loss $\mathcal{L}_{policy}$ such that its negative gradient approximates the reward gradient (i.e., $\nabla_\theta \mathcal{L}_{policy} \approx -\nabla_\theta \mathcal{J}$).We negate the approximate gradient derived above and add an $L_2$ regularization term for stability. This yields the final loss function used in Equation 17:$$\mathcal{L}_{policy}(\theta) = - \sum_{C_i} g_{C_i} \sum_{x \in C_i} \log v_\theta(x) + \beta \|\theta\|_2^2$$where $\beta$ is a hyperparameter controlling the strength of the regularization. By minimizing $\mathcal{L}_{policy}$, the optimizer increases the log-likelihood of selecting samples from clusters with positive advantages ($g_{C_i} > 0$) and suppresses those with negative impacts.

\section{Implementation Details}
We strictly control the computational cost to ensure fair comparisons. All models are trained with a fixed budget of 128M samples seen (e.g., $\sim$12 epochs for the full CC12M, or more epochs for smaller subsets).
\begin{figure}[h]
    \centering
    \includegraphics[width=1.0\textwidth]{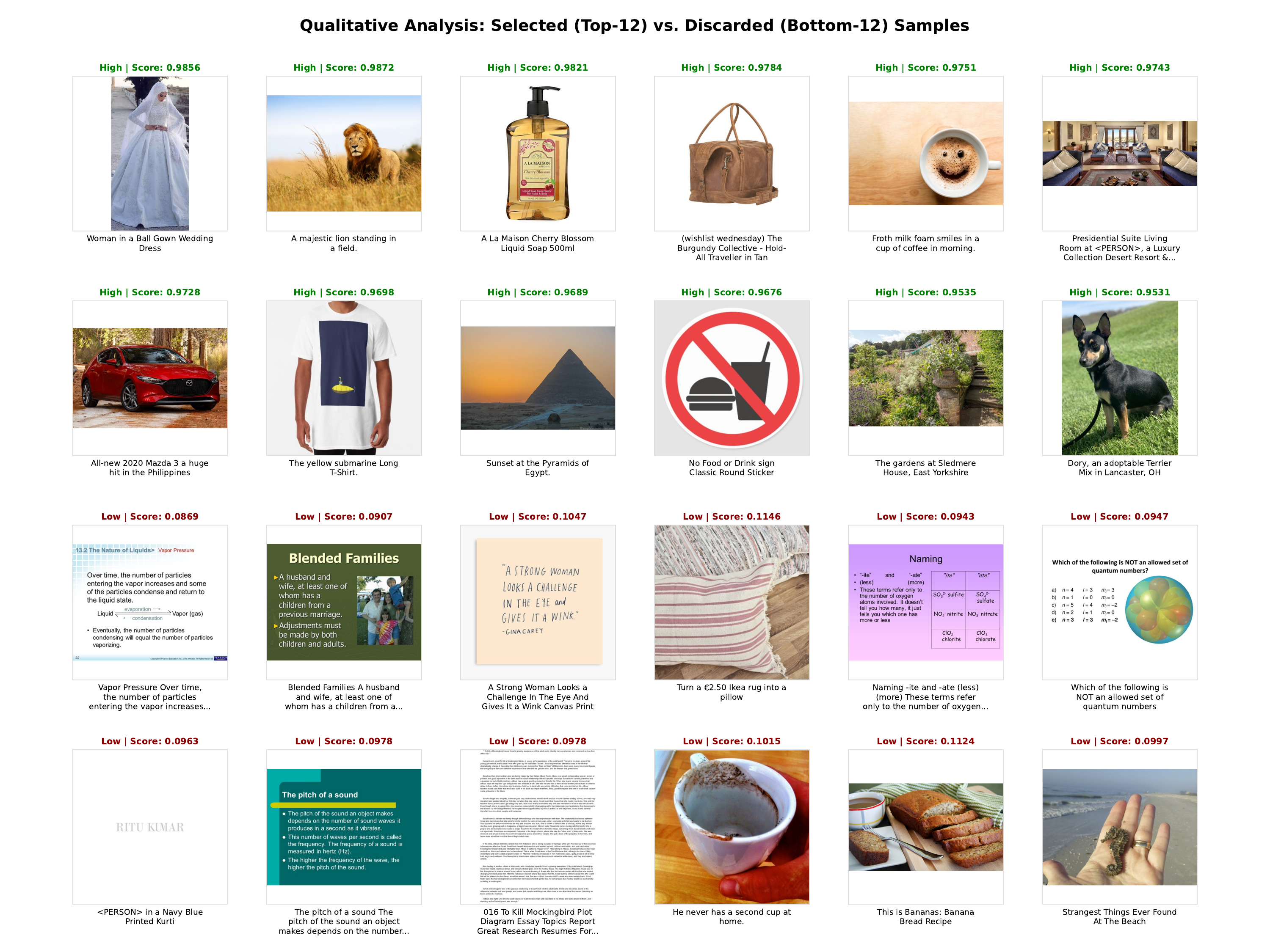}
    \caption{\textbf{Qualitative comparison of selected vs. discarded samples.} We visualize the samples with the highest (Top-12) and lowest (Bottom-12) scores. The high-scoring subset demonstrates strong semantic alignment, while the low-scoring subset filters out noise such as text-dominated slides and semantically irrelevant data.}
    \label{fig:qualitative_viz}
\end{figure}

\textbf{Proxy Model:} To provide efficient and faithful feedback for the FDO in TADS, we employ a ViT-B/16 CLIP proxy model initialized from OpenAI pre-trained weights, with all parameters fully trainable. The ViT-based proxy is adopted to better align the representation capacity and optimization dynamics with the final target model, leading to more reliable data valuation signals. The proxy model is optimized using the Adam optimizer with a learning rate of $5e^{-5}$ and a batch size of 4096, and is trained for 10 epochs, which empirically yields more stable and less noisy feedback without increasing the overall computational budget. 

\textbf{DVN:} The DVN is trained over 50 meta-iterations using Adam (learning rate $1e^{-3}$), with updates guided by downstream CLIP loss through the proxy model, ensuring the selection policy reflects task-specific data utility (batch size 1024). 

\textbf{Target Model:} The final selected subset is used to train a ViT-B/16 CLIP model from scratch. We use the Adam optimizer with a learning rate of $5e^{-4}$ and a batch size of 4096. All experiments are conducted on 8 NVIDIA A100 GPUs.

\section{Qualitative Analysis of Data Selection}
To explicitly visualize the efficacy of the proposed TADS framework, we present the top-12 highest-scoring and bottom-12 lowest-scoring samples ranked by our DVN in Figure~\ref{fig:qualitative_viz}. As observed, the High-Scoring Subset (Top-12) predominantly consists of high-resolution, object-centric images with captions that possess strong semantic alignment (e.g., clear product descriptions or scene descriptions). In contrast, the Low-Scoring Subset (Bottom-12) effectively identifies and discards low-utility noise, primarily comprising text-dominated presentation slides, abstract diagrams, and semantically irrelevant data where the visual content fails to match the textual description.

\section{Limitations and Discussion}
While TADS demonstrates strong empirical performance, several practical considerations and limitations remain.

\begin{itemize}
    \item \textbf{Sensitivity of Relevance Estimation in Low-Resource Regimes.}
    For downstream tasks characterized by sparse semantic definitions (i.e., extremely limited or abstract task prototypes), relevance estimation based on semantic similarity can be susceptible to spurious correlations inherent in the pre-trained embedding space. Although aggregating similarity scores across the task support set improves stability, the relevance vectors might still be driven by superficial matches—such as background textures or stylistic overlaps—rather than true task-discriminative semantics. Consequently, the selection policy might inadvertently favor samples that are visually congruent but semantically suboptimal. This limitation highlights the challenge of quantifying utility when the task definition itself is underspecified.

    \item \textbf{Error propagation from clustering in the deduplication stage.}
    In TADS, clustering is employed as part of the deduplication stage to identify and remove redundant or highly similar samples for improving data efficiency. However, imperfections in clustering may lead to inaccurate redundancy judgments, resulting in either false positives (removing non-redundant but informative samples) or false negatives (retaining redundant samples). Such errors can propagate to the subsequent distributional diversity weighting and gradient estimation stages. Specifically, mistakenly removed samples may reduce effective coverage of certain semantic regions, while retained redundant samples may distort density estimates and over-emphasize dominant patterns. These effects can alter sample contributions during optimization and potentially bias the learned data value function. While clustering-based deduplication significantly improves scalability, more robust and error-aware redundancy handling remains a promising direction for future work.


    \item \textbf{Adaptation to dynamic task sets.}
    When the downstream task set $\mathcal{M}$ changes, such as adding or removing tasks, the task relevance representations and the data valuation network (DVN) may require retraining to preserve effective multi-task alignment. This limitation arises from the coupling between data valuation and the task configuration during optimization. Developing lightweight or online adaptation mechanisms that can efficiently accommodate evolving task sets without full retraining remains an important direction for future work.

\end{itemize}


\end{document}